\documentclass[runningheads]{llncs}
\usepackage{graphicx}
\usepackage{algorithm}
\usepackage{algpseudocode}  
\usepackage{algorithmicx}
\usepackage{amsmath,color}
\renewcommand{\algorithmicreturn}{\textbf{Return} }
\algnewcommand\RETURN{\State \algorithmicreturn}%

\usepackage{multirow}
\begin{document}
\title{A Hybrid Evolutionary Algorithm for Reliable Facility Location Problem\thanks{This work was supported by the National Key R\&D Program of China (Grant No. 2017YFC0804003), the National Natural Science Foundation of China (Grant No. 61976111, 61906083), the Guangdong Provincial Key Laboratory (Grant No. 2020B121201001), the Program for Guangdong Introducing Innovative and Enterpreneurial Teams (Grant No. 2017ZT07X386), the Science and Technology Innovation Committee Foundation of Shenzhen (Grant No. JCYJ20190809121403553), the Shenzhen Science and Technology Program (Grant No. KQTD2016112514355531) and the Program for University Key Laboratory of Guangdong Province (Grant No. 2017KSYS008).The corresponding author: Xin Yao (xiny@sustech.edu.cn).}}
%
%
\author{Han Zhang\inst{}\orcidID{0000-0001-8243-1135} \and Jialin Liu\inst{}\orcidID{0000-0001-7047-8454} \and\\ Xin Yao\inst{}\orcidID{0000-0001-8837-4442}}
\authorrunning{H. Zhang, J. Liu, and X. Yao}
%
\institute{Guangdong Provincial Key Laboratory of Brain-inspired Intelligent Computation\\
Department of Computer Secience and Engineering\\ Southern University of Science and Technology\\ Shenzhen 518055, China.\\
\email{11849181@mail.sustech.edu.cn, \{liujl, xiny\}@sustech.edu.cn}\\
\url{} }
\maketitle              
\begin{abstract}
The reliable facility location problem (RFLP) is an important research topic of operational research and plays a vital role in the decision-making and management of modern supply chain and logistics. 
Through solving RFLP, the decision-maker can obtain reliable location decisions under the risk of facilities' disruptions or failures.
In this paper, we propose a novel model for the RFLP. Instead of assuming allocating a fixed number of facilities to each customer as in the existing works, we set the number of allocated facilities as an independent variable in our proposed model, which makes our model more close to the scenarios in real life but more difficult to be solved by traditional methods.
To handle it, we propose EAMLS, a hybrid evolutionary algorithm, which combines a memorable local search (MLS) method and an evolutionary algorithm (EA). Additionally, a novel metric called l3-value is proposed to assist the analysis of the algorithm's convergence speed and exam the process of evolution.  The experimental results show the effectiveness and superior performance of our EAMLS, compared to a CPLEX solver and a Genetic Algorithm (GA), on large-scale problems.


\keywords{reliable facility location problem  \and integer programming \and hybrid algorithm \and evolutionary algorithm \and local search.}
\end{abstract}
\section{Introduction}

The facility location problem aims at finding the optimal locations for facilities from a set of candidate location nodes  in order to minimize the cost such as the fixed facility cost and the transposition cost, or to maximize the total revenue. In general, there are also some constraints to be considered, such as satisfying all customers' demands, etc. It is an NP-hard optimization problem \cite{ref5,ref6,ref7} and has attracted much attention from researchers in both the scientific community and engineering field due to its wide application in real world. The facilities could be hospitals, restaurants, post stations, bus stations, industrial plants, banks, warehouses, and distribution centers, etc. The facility location decision has high precedence in the whole logistics decisions and has a great influence on subsequent operation level decisions \cite{ref4}. Daskin et al. \cite{ref5} regards the location decisions as “the most critical and most difficult of the decisions needed to realize an efficient supply chain”.

In RFLP, the facility is not always available all the time \cite{ref5}. One or more of them may not work from time to time because of disruptions, examples include natural disasters, inclement weather, destruction of facilities by fire or flood, expiration of the contract, and any other force majeure factors. In such a situation, these are facility “failures”. The failures of the facilities will result in excessive transportation costs because the customers that were considered to be served by them must be served by other, usually more distant, facilities \cite{ref5}. Therefore, by solving RFLP, we can get a location decision which can ensure a certain level of reliability to guarantee customers can get service when facilities' failures occur.

Many models have been proposed for RFLP, in which all kinds of factors were taken into account and many of them are formulated for specific applications in real life. In addition, large-scale RFLP problems have rarely been considered. The algorithms studied in literature were mainly tested on problems of small size.

This paper focuses on two aspects: the problem formulation and the algorithm. Based on the work of~\cite{ref8,ref9}, we propose a new reliable facility location-allocation problem (RFLP) formulation, which does not fix the number of allocated facilities to each customer as a constant and is more close to reality. The resulted model is a nonlinear 0-1 integer programming model which is more complicated for traditional methods. In this paper, a hybrid evolutionary algorithm called EAMLS is proposed to solve it. EAMLS combines a memorable local search method with an evolutionary algorithm, which has a good performance on both small-scale and large-scale problems considered in this paper. It is worth mentioning that the instances used in our experiments are much larger than the ones used in previous work. Furthermore, a convergence metric l3-value is proposed for analyzing the algorithm and observing the evolutionary process.

The rest of this paper is organized as follows. Section 2 briefly reviews the related work of RFLP. In Section 3, our new RFLP formulation is introduced. We proposed a hybrid evolutionary algorithm EAMLS in Section 4. Section 5 presents computational studies, and Section 6 concludes.

\section{Related Work}

By solving a specific RFLP, decision-makers expect to get a robust location decision which is still economical when some facilities fail under various disruptions. The research can be divided into two categories according to the method used to handle facility failure or ensure reliability.

Some works \cite{ref10,ref11,ref12} use a disruptive scenarios approach to describe facility failure. In this approach, scenarios contain facility failure information, e.g., simultaneously disrupted facility sites, modified customer demands, and facility costs, etc. The disruptive scenarios approach can describe the facility failure information well, but it usually requires plenty of scenarios to cover different disruptive situations, which implies large computational cost, especially for large-scale problems.

Another approach to ensure reliability is to allocate two or more facilities to serve each customer \cite{ref8,ref9,ref16,ref17}. In this approach, the method for reliability is intuitive and easy to understand. Both a location decision (which contains how many facilities needed to build and where to build them) and an allocation decision (which shows how to allocate facilities to serve customers ) are determined before the occurrences of facilities' disruptions/failures.

Some RFLP models have been proposed, e.g., models proposed by Li et al. \cite{ref8} and Snyder and Darskin \cite{ref9}. Table \ref{tab:notations} summarizes the notations used in the models. 

\begin{table}[htbp]
\centering
\caption{Description of notations.}
\label{tab:notations}
\resizebox{\textwidth}{!}{%
\begin{tabular}{|c|c|c|c|}
\hline
Notations & Description                                             & Notations & Description                                   \\ \hline
$I$       & the set of customers, index by $i$;                     & $m$       & \# of facilities allocated for each customer; \\
$J$      & the set of candidate location sites. index by $j$;   & $p$   & the facility failure probability; \\
$NF$      & the set of candidate location sites that will not fail; & $f_i$     & the fix cost of $j$;                          \\
$F$       & the set of candidate location sites that may fail;      & $\alpha$  & weighted parameter;                           \\
$c_{ij}$ & the cost of per unit demand shipped from $j$ to $i$; & $h_i$ & the demands of customer $i$;       \\ \hline
\end{tabular}%
}
\end{table}

\noindent Besides, there are two sets of decision variables: location decision variables ($\mathbf{X}$) and allocation decision variables ($\mathbf{Y}$):
\begin{equation}
    \label{variable:X}
    X_j=\left\{\begin{array}{lc}1,&\text{if candidate location site}\;j\;\text{is selected};\\0,&\text{otherwise}.\end{array}\right.
\end{equation}
\begin{equation}
    \label{variable:Y}
    Y_{ijr}=\left\{\begin{array}{lc}1,&\text{if } j \text{ is allocated as  the level-} r \text{ facility to serve } i;\\0,&\text{otherwise}.\end{array}\right.
\end{equation}

\noindent In Eq. (\ref{variable:Y}), the "level-r" facility $j$ for customer $i$ means the facility $j$ will provide service only when the front $r$ allocated facilities (from level-0 to level-(r-1)) fail.

A classical RFLP model in \cite{ref9} is as follows.

\begin{equation}
    \label{objfun2-1}
    \operatorname{Min} \alpha w_{1}+(1-\alpha) w_{2}
\end{equation}
Subject to:
\begin{equation}
    \label{const2-2}
    w_{1}=\sum_{j \in J} f_{j} X_{j}+\sum_{i \in I} \sum_{j \in J} h_{i} c_{i j} Y_{i j 0}
\end{equation}
\begin{equation}
    \label{const2-3}
    w_{2}=\sum_{i \in I} h_{i}\left[\sum_{j \in N F} \sum_{r=0}^{m-1} c_{i j} p^{r} Y_{i j r}+\sum_{j \in F} \sum_{r=0}^{m-1} c_{i j} p^{r}(1-p) Y_{i j r}\right]
\end{equation}
\begin{equation}
    \label{const2-4}
    \sum_{j \in J} Y_{i j r}+\sum_{j \in N F} \sum_{t=0}^{r-1} Y_{i j t}=1 \quad \forall i \in I, r=0, \ldots, m-1
\end{equation}
\begin{equation}
    \label{const2-5}
    Y_{i j r} \leq X_{j} \quad \forall i \in I, j \in J, r=0, \ldots, m-1
\end{equation}
\begin{equation}
    \label{const2-6}
    \sum_{r=0}^{m-1} Y_{i j r} \leq 1 \quad \forall i \in I, \forall j \in J
\end{equation}
\begin{equation}
    \label{const2-7}
    m=|J|
\end{equation}
\begin{equation}
    \label{const2-8}
    X_{u}=1
\end{equation}
\begin{equation}
    \label{const2-9}
    X_{j} \in\{0,1\} \quad \forall j \in J
\end{equation}
\begin{equation}
    \label{const2-10}
    Y_{ijr} \in\{0,1\} \quad \forall i \in I; \forall j \in J; r=0, \ldots, m-1
\end{equation}

\noindent In this model, there are two objectives in the objective function, $w_1$ is the operating cost and $w_2$ is the expected failure cost. The objective of the model is to minimize the weighted sum of the two objectives. 
Besides, there is an emergency facility $u$ which will always be selected and not fail, and all customers can get service from it.

Several shortcomings are observed in the literature:

(1) The number of facilities allocated to each customer (i.e., $m$ in Eq. (\ref{const2-7})) is fixed in models of most literature, e.g., $m = 2$ (i.e., $Y_{ij0}$ and $Y_{ij1}$) in \cite{ref8} and $m = |J|$ in \cite{ref9}. One issue of this allocation setting is the determination of an appropriate value of $m$. If $m$ is bigger than the number of selected candidate location sites, i.e., $\sum\nolimits_{j\in J}X_j$, it is not in line with the actual situation because we cannot allocate 
nonexistent facilities to customers. If we set the value of $m$ smaller than $\sum\nolimits_{j\in J}X_j$, the value of $\sum\nolimits_{j\in J}X_j$ is changed during the exploration in solution space, therefore it is hard for us to set a suitable $m$ value. If we set $m = 2$ directly, which means allocate just one primary facility and one backup facility to serve each customer, the reliability is a bit weak intuitively.

(2) To our best knowledge, there is a lack of research on the large-scale problem. The largest problem instance in the related research is 150-node and the optimization solver such as CPLEX can find near-optimal or even optimal solutions for the problem.

(3) There is a lack of research on the algorithm which can solve the large-scale problems efficiently as well.

Correspondingly, this paper:

(1) constructs a new formulation in which a non-fixed allocation setting, i.e., $m=\sum_{j \in J}X_j$, is used;

(2) proposes a hybrid evolutionary algorithm EAMLS which combines a local search method with an evolutionary algorithm and performs well on both small-scale and large-scale problems;

(3) performs experimental studies on large-scale problems whose scale is much larger than any related literature;

(4) proposes a convergence metric l3-value to help observe the evolutionary process, adjust parameters and further improve the algorithm.

\section{Problem Formulation}

We propose a new RFLP formulation in which we set the number of allocated facilities to each customer as an variable instead of a fixed constant.

The mathematical formulation of our model is as follows, formulated based on \cite{ref8,ref9}. The decision variables are defined by Eqs. (\ref{variable:X}) and (\ref{variable:Y}).

\begin{equation}
    \label{objectiveFunction:3-1}
    \operatorname{Min} \sum_{j \in J} f_{j} X_{j}+\alpha \sum_{i \in I} \sum_{j \in J} \sum_{r=0}^{m-1} h_{i} c_{i j} p^{r}(1-p) Y_{i j r}
\end{equation}
Subject to:
\begin{equation}
    \label{constraint3-5}
    m=\sum_{j \in J} X_{j}
\end{equation}
\begin{equation}
    \label{constraint3-4}
    m \geq 2
\end{equation}
\begin{equation}
    \label{constraint:3-2}
    \sum_{j \in J} Y_{i j r}=1 \quad \forall i \in I ; r=0, \ldots, m-1
\end{equation}
\begin{equation}
    \label{constraint3-3}
    \sum_{r=0}^{m-1} Y_{i j r} \leq X_{j} \quad \forall i \in I, \forall j \in J
\end{equation}
\begin{equation}
    \label{constraint3-6}
    X_{j} \in\{0,1\} \quad \forall j \in J
\end{equation}
\begin{equation}
    \label{constraint3-7}
    Y_{ijr} \in\{0,1\} \quad \forall i \in I; \forall j \in J; r=0, \ldots, m-1
\end{equation}

\noindent The objective function of the model is to minimize the total cost associate with facilities construction (i.e., the term $\sum_{j \in J}f_jX_j$) and transportation between the facilities and customers (i.e., the term $\sum_{i \in I}\sum_{j \in J}\sum_{r=0}^{m-1}h_ic_{ij}p^r(1-p)Y_{ijr}$).

Constraint (\ref{constraint3-5}) makes the number of facilities allocated to each customer (i.e., $m$) a variable and its value is related to location decision variables (i.e., $\mathbf{X}$). Constraint (\ref{constraint3-4}) represents at lease two facilities are constructed to ensure reliability. Constraint (\ref{constraint:3-2}) assures only one facility can be the level-$r$ supplier of customer $i$. Constraint (\ref{constraint3-3}) means candidate location site $j$ can be allocated to customer as a supplier only when it is selected.  Constraint (\ref{constraint3-6}) and (\ref{constraint3-7}) are standard integrality constraints.

Compared with classical models shown in Section 2, the significant difference in our model is the new non-fixed facility allocation setting, i.e., constraint (\ref{constraint3-5}). In our model, the value of $m$ is not fixed but varies with decision variables $\mathbf{X}$, therefore it is more realistic, ensures reliability, but makes our model much more complex and difficult to solve by traditional methods as well.

\section{A Hybrid Evolutionary Algorithm: EAMLS}

This paper develops a new hybrid evolutionary algorithm EAMLS (Evolutionary Algorithm with Memorable Local Search) which combines a memorable local search method and an EA, and a convergence metric l3-value is proposed. In this section, the structure of EAMLS is explained first, then the design of operators of the Genetic Algorithm (GA) and EAMLS is introduced. Finally, the details of l3-value are described.

\subsection{EAMLS}

Algorithm \ref{algorithm1} is the pseudo-code of EAMLS. Compared with the GA, the main characters of EAMLS contain: (1) no crossover operation; (2) population size self-adaptation; (3) the combination of a memorable local search (MLS) and EA; and (4) the adoption of convergence metric l3-value.

In Algorithm \ref{algorithm1}, variable $allNeighborInds$ stores all non-repeating neighborhood individuals generated by MLS before current generation and is updated at the end of every generation (Algorithm 1, Line 2 and Line 13). In the evolutionary process, a new population is generated from the current population after mutation, MLS, and survival selection (Algorithm 1, Lines 5-8), and convergence metric l3-value is calculated (Algorithm 1, Line 9). If l3-value is bigger than a pre-set threshold $\beta$, population size is increased by a pre-set step size $p$ (Algorithm 1, Lines 10-12). The description of the l3-value will be shown in Section 4.3.

\begin{algorithm}[htbp]
  \caption{Evolutionary Algorithm with Memorable Local Search.}
  \label{algorithm1}
  \begin{algorithmic}[1]
    \Require
      $G$: number of generations;
      $\mu$: population size;
      $l$: individual length;
      $m$: mutation rate;
      $\beta$: threshold of l3-value;
      $p$: step size of population self-adaptation;
    \Ensure
      $bestSol$: the best individual in the final population;
    \State $initPop \leftarrow initializePop(\mu,l);$
    \State $allNeighborInds \leftarrow$ an empty set;
    \State $pop \leftarrow evaluatePop(initPop);$
    \For{$g=1 \ to\ G$}
        \State $popAfterMutation \leftarrow mutation(pop,m);$
        \State $offspring \leftarrow evaluatePop(popAfterMuation);$
        \State $offspring_{LS} \leftarrow memorableLocalSearch(pop,offspring);$
        \State $pop \leftarrow survival(pop,offspring,offspring_{LS},\mu);$
        \State l3-value $\leftarrow getl3Value(pop,allNeighborInds);$
        \If{l3-value$ > \beta$}
            \State $\mu \leftarrow \mu+p;$
        \EndIf
        \State add $ offspring_{LS}$ to $allNeighborInds;$
    \EndFor
    \State $bestSol \leftarrow selectBestIndividual(pop)$
    \RETURN{} $bestSol$
  \end{algorithmic}
\end{algorithm}

\begin{algorithm}[htbp]
  \caption{Memorable Local Search.}
  \label{algorithm2}
  \begin{algorithmic}[1]
    \Require
      $pop$: the parent population;
      $offspring$: the child population generated after mutation;
      $n$: \# of individuals which need to check whether to do local search;
      $indLSed$:the set of individuals which have already down local search before this generation;
    \Ensure
      $offspring_{LS}$: the population generated by local search;
    \State $offspring_{LS}\leftarrow$ an empty set;
    \State $parentPop \leftarrow$ combine $pop$ and $offspring$;
    \State $sortedParentPop\leftarrow$ sort $parentPop$ by fitness increasing order;
    \State $i \leftarrow 0;$
    \For{$j \leftarrow 1$ to $len(sortedParentPop)$}
        \If{$sortedParentPop[j]$ not in $indLSed$}
            \State $neighborInds \leftarrow generateNeighbor(sortedParentPop[j]);$
            \State add $neighborInds$ to $offspring_{LS};$
            \State $i \leftarrow i+1;$
            \If{$i > n$}
                \State break;
            \EndIf
        \EndIf
    \EndFor
    \RETURN $offspring_{LS}$
  \end{algorithmic}
\end{algorithm}

Algorithm \ref{algorithm2} is the pseudo-code of the memorable local search (MLS). First, we will introduce the definition of the neighborhood. The neighborhood of an individual is the set of individuals whose Hamming distance is 1 from that individual. In MLS, sort $(\mu+\lambda)$ population (variable $sortedParentPop$ in Algorithm 2) in decreasing order, i.e., good individuals are in the front. Then check individuals one by one in sorted $(\mu+\lambda)$ population whether it has been local-searched before this generation, and do local-search for those have not been local-searched (Lines 5-7 in Algorithm 2. 
It looks like that the algorithm remembers all local-searched individuals and that's why we name it 
Memorable Local Search). Exit the loop until the number of new individuals which have been local-searched in this generation reaches $n$ (Lines 9-12 in Algorithm 2).

\subsection{Operator Design of GA and EAMLS}

In Section 5, we use a GA for comparison. Here some operators' design for GA and EAMLS is as follows \footnote{If there is no special statement, that operator is adopted in both GA and EAMLS.}:

\textbf{Representation} This paper uses binary representation. Every bit represents a location decision variable $X_j, j \in J$.

\textbf{Population Initialization} Stochastic initialization is used in GA and EAMLS. Every gene of an individual takes 0 or 1 with equal probability.

\textbf{Fitness Function} In general, the bigger the fitness value is, the better the individual will be. Therefore, the reciprocal of the objective value of the individual is used as the fitness function.

\textbf{Selection Operator} In GA, roulette wheel selection is used to select parents to do crossover operation.

\textbf{Crossover Operator} In GA, a one-point crossover operator is used. For two parent individuals selected by the selection operator, do crossover operation according to a pre-set crossover rate.

\textbf{Mutation Operator} The bit-flipping mutation is used in GA and EAMLS. During mutation, every gene/bit of one individual mutates with a pre-set mutation rate.

\textbf{Survival Selection Strategy} We adopt $(\mu+\lambda)$ strategy to select next generation population from $(\mu+\lambda)$ population, i.e., the mixed population of the current generation population and the offspring.

\textbf{Repair Strategy} Repair strategy is working when there are individuals which do not satisfy the constraint (\ref{constraint3-4}). For an individual needed repair, check every gene in ascending order of fixed cost and change the gene with 0-value to 1 until the individual satisfies the constraint (\ref{constraint3-4}).

\textbf{How to determine $\mathbf{Y}$} For one customer, the selected candidate locations (i.e., locations whose $X_j = 1$) are allocated to it in ascending order of distance, which has been proved the optimal allocation pattern under a certain solution $\mathbf{X}$ \cite{ref9} and can satisfy the constraints (12), (13), and (15).

\subsection{Convergence Metric l3-value}

In order to observe the evolutionary process, a convergence metric l3-value is proposed.

Algorithm \ref{algorithm3} is the pseudo-code of the calculation method of l3-value. The new population generated after survival selection is checked, and the number of individuals which also belong to the set $allNeighborInds$ is counted (Lines 2-6 in Algorithm 3). Then we calculate the proportion of these individuals in the population as l3-value (Line 7 in Algorithm \ref{algorithm3}).
l3-value can be used to measure the convergence during the evolutionary process. The bigger the l3-value is, the stronger the evolution converges.

\begin{algorithm}[htbp]
  \caption{Function $getl3Value()$.}
  \label{algorithm3}
  \begin{algorithmic}[1]
    \Require
    $pop$: the new population after survival selection;
    $allNeighborInds$: the set of all individuals generated by memorable local search before this generation;
    \Ensure
     l3-value;
    \State $num\leftarrow 0;$
    \For{$ind \in pop$}
        \If{$ind \in allNeighborInds$}
            \State $num \leftarrow num+1;$
        \EndIf
    \EndFor
    \State l3-value $\leftarrow num/len(pop);$
    \State{\textbf{Return} $l3$-$value$} 
  \end{algorithmic}
\end{algorithm}

\section{Computational Studies}

Because this paper proposes a new problem, and there are not any algorithms like EAMLS can be used to compare directly, we compare EAMLS with a GA and CPLEX (a commercial optimization solver of IBM) on two models: m=2 and m=${\textstyle\sum_{j\in J}}X_j$ models. The difference between the two models is the allocation setting. In the m=2 model, the number of facilities allocated to each customer, i.e. $m$, is fixed to 2, which is adopted in much literature. The m=${\textstyle\sum_{j\in J}}X_j$ model is proposed by us in this paper and $m$ varies with decision variables $\mathbf{X}$ during the search process. Section 5.1 shows the experimental design, including instances generation, parameters setting, and experimental environment. The experiments and results of the m=2 and m=${\textstyle\sum_{j\in J}}X_j$ models are presented in Sections 5.2. Analyses and discussions are given in Section 5.3.

\subsection{Experimental Design}

\textbf{Instance Generation} This paper generates problem instances uniformly at random on different scales. The parameters used to generate instances are shown in Table \ref{tab:5-1tab2}. There are eight 10-node instances, eight 50-node instances, eight 100-node instances, and four 600-node instances.

\begin{table}[]
\centering
\caption{Parameters used in instances generation}
\label{tab:5-1tab2}
\begin{tabular}{|c|c|}
\hline
Parameters                    & Ranges               \\ \hline
Candidate location coordinate & {[}0,1{]}            \\
Customer demands              & \{0,1,...,1000\}     \\
Fixed cost of facility        & \{500,501,...,1500\} \\
Facility failure probability  & 0.05                 \\ \hline
\end{tabular}
\end{table}

\noindent\textbf{Parameter Setting of Algorithms} Some parameters’ values of GA and EAMLS are shown in Table \ref{tab:5-2tab3}. Table \ref{tab:5-3tab4} presents the generation number and population size of GA and EAMLS, which associate with the scale of problem instances.
The values of parameters in Tables \ref{tab:5-2tab3} and \ref{tab:5-3tab4} are chosen arbitrarily on the basis of meeting the following conditions: (1) EAMLS converges at the end of evolution; (2) the number of fitness evaluations (FEs) of GA is not lower than EAMLS. Besides, the default parameters of CPLEX are used.

\begin{table}[]
\centering
\caption{Some parameters of GA and EAMLS}
\label{tab:5-2tab3}
\begin{tabular}{|c|c|}
\hline
Parameters                                & Value \\ \hline
Crossover rate for GA, $c$                        & 0.9   \\
Mutation rate, $m$                         & 0.1   \\
\# Local search individual, $n$            & 10    \\
l3-value threshold, $\beta$                & 0.8   \\
Step size of population self-adaption, $p$ & 100   \\ \hline
\end{tabular}
\end{table}

\begin{table}[]
\centering
\caption{Parameters associate with instance size}
\label{tab:5-3tab4}
\begin{tabular}{|c|c|c|c|c|}
\hline
\multirow{2}{*}{\begin{tabular}[c]{@{}c@{}}Instance scale\\ (\# nodes)\end{tabular}} & \multicolumn{2}{c|}{GA}         & \multicolumn{2}{c|}{EAMLS}      \\ \cline{2-5} 
                                                                                     & \# Generation & Population size & \# Generation & Population size \\ \hline
10  & 60   & 30  & 10  & 20  \\
50  & 200  & 200 & 20  & 20  \\
100 & 400  & 200 & 50  & 100 \\
600 & 4600 & 200 & 250 & 200 \\ \hline
\end{tabular}
\end{table}

\textbf{Experimental Environment} The algorithms are implemented in Python 3.7 and run on Dell R370 server which has 2x Intel(R) Xeon(R) CPU E5-2650 v4 @ 2.20GHz CPU, 128G RAM, and CentOS 7.6 operating system.

\textbf{Statistical Test} We use the Wilcoxon sign rank test to determine whether the results between EAMLS and other methods have statistically significant differences. The Wilcoxon sign rank test is a non-parameter test which is suitable for two related or matched samples and compares data in pair, hence it is suitable to use here.

\subsection{Experiments on the m=2 and m=${\textstyle\sum_{j\in J}}X_j$ Models}

For the m=2 model, We compare EAMLS with the GA and CPLEX on small-scale (10-node), mid-scale (100-node), and large-scale (600-node) instances. There are 30 runs on small and mid-scale instances and 10 runs on large-scale instances because of time. The computational results are shown in Table \ref{tab:mIs2_10-100-600-nodetab5}.

\begin{table}[]
\centering
\caption{Computational results on m=2 model 10 (30 runs),100 (30 runs), and 600 (10 runs)-node instances. AOV is Average Objective Value. OR is the Optimal Rate and calculated by (\# runs which finding the optimal solution)/(\# all runs). Gap is calculated by (AOV(other method)-AOV(EAMLS))/AOV(EAMLS). When Gap is positive, the performance of other methods is worse than EAMLS, otherwise better. The symbol “*” in AOV represents the results between EAMLS and that method have statistically significant differences. The symbol “-” represents CPLEX cannot solve the instance or the optimal solution is unknown so no results can be given.}
\label{tab:mIs2_10-100-600-nodetab5}
\resizebox{\textwidth}{!}{%
\begin{tabular}{|c|c|c|c|c|c|c|c|c|c|c|c|c|}
\hline
\multirow{2}{*}{\begin{tabular}[c]{@{}c@{}}Instance\\ No.\end{tabular}} & \multicolumn{4}{c|}{GA}                                 & \multicolumn{4}{c|}{CPLEX}                           & \multicolumn{4}{c|}{EAMLS}                               \\ \cline{2-13} 
                                                                        & AOV              & Gap (\%) & OR            & Time      & AOV               & Gap (\%) & OR   & Time           & AOV               & Gap (\%) & OR   & Time               \\ \hline
10-1                                                                    & \textbf{2463.19} & 0.00     & \textbf{1.00} & 1.52      & \textbf{2463.19}  & 0.00     & 1.00 & \textbf{0.46}  & \textbf{2463.19}  & 0.00     & 1.00 & 6.14               \\
10-2                                                                    & \textbf{2874.03} & 0.00     & \textbf{1.00} & 1.51      & \textbf{2874.03}  & 0.00     & 1.00 & \textbf{0.41}  & \textbf{2874.03}  & 0.00     & 1.00 & 5.46               \\
10-3                                                                    & \textbf{2623.35} & 0.00     & \textbf{1.00} & 1.74      & \textbf{2623.35}  & 0.00     & 1.00 & \textbf{0.66}  & \textbf{2623.35}  & 0.00     & 1.00 & 5.41               \\
10-4                                                                    & \textbf{2323.92} & 0.00     & \textbf{1.00} & 1.93      & \textbf{2323.92}  & 0.00     & 1.00 & \textbf{0.48}  & \textbf{2323.92}  & 0.00     & 1.00 & 5.86               \\
10-5                                                                    & \textbf{2917.87} & 0.00     & \textbf{1.00} & 2.48      & \textbf{2917.87}  & 0.00     & 1.00 & \textbf{0.50}  & \textbf{2917.87}  & 0.00     & 1.00 & 5.71               \\
10-6                                                                    & \textbf{3149.31} & 0.00     & \textbf{1.00} & 2.72      & \textbf{3149.31}  & 0.00     & 1.00 & \textbf{0.41}  & \textbf{3149.31}  & 0.00     & 1.00 & 5.59               \\
10-7                                                                    & \textbf{3324.98} & 0.00     & \textbf{1.00} & 2.39      & \textbf{3324.98}  & 0.00     & 1.00 & \textbf{0.58}  & \textbf{3324.98}  & 0.00     & 1.00 & 5.64               \\
10-8                                                                    & \textbf{3165.87} & 0.00     & \textbf{1.00} & 2.10      & \textbf{3165.87}  & 0.00     & 1.00 & \textbf{0.52}  & \textbf{3165.87}  & 0.00     & 1.00 & 4.58               \\
100-1                                                                   & 13029.83*        & 22.28    & 0.00          & 2374.86   & \textbf{10645.89} & -0.10    & 1.00 & \textbf{14.30} & 10656.11          & 0.00     & 0.87 & 1431.17            \\
100-2                                                                   & 13166.44*        & 20.95    & 0.00          & 2375.71   & \textbf{10885.43} & 0.00     & 1.00 & \textbf{14.31} & \textbf{10885.43} & 0.00     & 1.00 & 1387.01            \\
100-3                                                                   & 12982.37*        & 16.90    & 0.00          & 2396.42   & \textbf{11105.21} & 0.00     & 1.00 & \textbf{14.76} & 11105.39          & 0.00     & 0.93 & 1514.12            \\
100-4                                                                   & 13379.41*        & 16.66    & 0.00          & 2388.67   & \textbf{11468.64} & 0.00     & 1.00 & \textbf{14.42} & \textbf{11468.64} & 0.00     & 1.00 & 1382.91            \\
100-5                                                                   & 14563.46*        & 16.39    & 0.00          & 2398.34   & \textbf{12505.51} & -0.05    & 1.00 & \textbf{14.80} & 12512.29          & 0.00     & 0.90 & 1415.63            \\
100-6                                                                   & 13189.74*        & 17.29    & 0.00          & 2402.44   & \textbf{11245.55} & 0.00     & 1.00 & \textbf{14.00} & \textbf{11245.55} & 0.00     & 1.00 & 1447.11            \\
100-7                                                                   & 12841.37*        & 16.11    & 0.00          & 1696.85   & \textbf{11043.70} & -0.15    & 1.00 & \textbf{15.49} & 11059.89          & 0.00     & 0.90 & 1326.41            \\
100-8                                                                   & 13886.78*        & 18.30    & 0.00          & 1242.25   & \textbf{11732.46} & -0.05    & 1.00 & \textbf{14.94} & 11738.83          & 0.00     & 0.87 & 1180.91            \\
600-1                                                                   & 144896.91*       & 281.04   & -             & 655420.67 & -                 & -        & -    & -              & \textbf{38026.65} & 0.00     & -    & \textbf{564432.00} \\
600-2                                                                   & 145508.23*       & 293.12   & -             & 656832.50 & -                 & -        & -    & -              & \textbf{37013.71} & 0.00     & -    & \textbf{572568.34} \\
600-3                                                                   & 141486.28*       & 283.41   & -             & 654632.01 & -                 & -        & -    & -              & \textbf{36902.36} & 0.00     & -    & \textbf{568824.96} \\
600-4                                                                   & 141256.35*       & 282.80   & -             & 656656.21 & -                 & -        & -    & -              & \textbf{36900.52} & 0.00     & -    & \textbf{568546.96} \\ \hline
\end{tabular}%
}
\end{table}

For the m=${\textstyle\sum_{j\in J}}X_j$ model, we compare EAMLS with the GA and CPLEX on 50 and 100-node instances, and there are 30 runs on each instance. Table \ref{tab:mIsAll_50100-node} is the computational results.

\begin{table}[]
\centering
\caption{Computational results on m=$\sum_{j \in J}X_j$ model 50 and 100-node instances, 30 runs. AOV is Average Objective Value. Gap is calculated by ((AOV(other method)-AOV(EAMLS))/AOV(EAMLS). When Gap is positive, the performance of other methods is worse than EAMLS, otherwise better. The symbol “*” in AOV represents the results between EAMLS and that method have statistically significant differences.}
\label{tab:mIsAll_50100-node}
\resizebox{\textwidth}{!}{%
\begin{tabular}{|c|c|c|c|c|c|c|c|c|c|}
\hline
\multirow{2}{*}{\begin{tabular}[c]{@{}c@{}}Instance\\ No.\end{tabular}} & \multicolumn{3}{c|}{GA}        & \multicolumn{3}{c|}{CPLEX}       & \multicolumn{3}{c|}{EAMLS}                      \\ \cline{2-10} 
                                                                        & AOV       & Gap (\%) & Time    & AOV        & Gap (\%) & Time     & AOV               & Gap (\%) & Time             \\ \hline
50-1                                                                    & 7053.71*  & 0.68     & 719.76  & 12589.41*  & 79.69    & 4715.56  & \textbf{7006.23}  & 0.00     & \textbf{91.52}   \\
50-2                                                                    & 7154.93   & -0.13    & 720.49  & 15734.80*  & 119.63   & 4488.26  & \textbf{7164.20}  & 0.00     & \textbf{90.75}   \\
50-3                                                                    & 6890.54*  & 0.75     & 713.25  & 12656.13*  & 85.06    & 5219.33  & \textbf{6838.95}  & 0.00     & \textbf{91.10}   \\
50-4                                                                    & 7166.63   & 0.04     & 698.45  & 12147.92*  & 69.58    & 4702.01  & \textbf{7163.42}  & 0.00     & \textbf{90.04}   \\
50-5                                                                    & 6929.29   & 0.03     & 714.86  & 11946.35*  & 72.46    & 5281.27  & \textbf{6926.95}  & 0.00     & \textbf{87.42}   \\
50-6                                                                    & 6575.09   & 0.29     & 696.80  & 13284.69*  & 102.64   & 4836.45  & \textbf{6555.87}  & 0.00     & \textbf{90.59}   \\
50-7                                                                    & 7162.83   & 0.07     & 685.04  & 12441.00*  & 73.81    & 4495.41  & \textbf{7157.76}  & 0.00     & \textbf{81.19}   \\
50-8                                                                    & 7175.89*  & 0.26     & 629.19  & 14433.41*  & 101.67   & 4522.35  & \textbf{7156.99}  & 0.00     & \textbf{70.07}   \\
100-1                                                                   & 12895.10* & 20.47    & 3976.97 & 113781.45* & 963.00   & 19451.62 & \textbf{10703.78} & 0.00     & \textbf{2266.50} \\
100-2                                                                   & 13093.80* & 19.77    & 3820.11 & 110441.89* & 910.18   & 17159.57 & \textbf{10932.89} & 0.00     & \textbf{2168.54} \\
100-3                                                                   & 13082.38* & 17.21    & 2719.68 & 114576.21* & 926.52   & 35836.91 & \textbf{11161.59} & 0.00     & \textbf{2337.35} \\
100-4                                                                   & 13484.69* & 17.04    & 2551.55 & 99484.65*  & 763.50   & 35129.74 & \textbf{11521.11} & 0.00     & \textbf{2217.00} \\
100-5                                                                   & 14484.70* & 15.22    & 2579.12 & 111338.15* & 785.68   & 17249.10 & \textbf{12570.86} & 0.00     & \textbf{2279.35} \\
100-6                                                                   & 13360.41* & 18.20    & 2626.97 & 99397.46*  & 779.39   & 19433.87 & \textbf{11302.96} & 0.00     & \textbf{2288.82} \\
100-7                                                                   & 12810.60* & 15.20    & 2553.83 & 105460.22* & 848.32   & 17271.95 & \textbf{11120.78} & 0.00     & \textbf{2036.58} \\
100-8                                                                   & 13809.12* & 17.05    & 2548.32 & 112170.76* & 850.82   & 18667.88 & \textbf{11797.25} & 0.00     & \textbf{1792.08} \\ \hline
\end{tabular}%
}
\end{table}

\subsection{Analyses and Discussions}

We compare GA, CPLEX, and EAMLS on different scale (10, 100, and 600-node) problem instances for m=2 model whose allocation setting is often used in literature, and the experimental results are shown in Table \ref{tab:mIs2_10-100-600-nodetab5}. Experimental results on 50 and 100-node instances of the new complicated m=${\textstyle\sum_{j\in J}}X_j$ model are presented in Table \ref{tab:mIsAll_50100-node}. 

For m=2 model, from Table \ref{tab:mIs2_10-100-600-nodetab5}, we can see that CPLEX performs the best on both solution quality and time for small and mid-scale (10 and 100-node) instances. EAMLS can find solutions as good as CPLEX but need more time. Although CPLEX can solve small and mid-scale instances fast, it needs more RAM space as the problem scale increases. For large-scale problem (600-node) instances, EAMLS can find better solutions in less time compared with GA, while the CPLEX cannot find a solution.

The new m=${\textstyle\sum_{j\in J}}X_j$ model is more complicated to solve, especially for CPLEX. Table \ref{tab:mIsAll_50100-node} demonstrates that the performance of EAMLS is better than GA and CPLEX on both solution quality and time.

According to the observation of computational results, we can get three features of EAMLS: (1) For small- and mid-scale problems, the solutions found by EAMLS are comparable to those found by other methods; (2) For large-scale problems, EAMLS significantly outperforms other methods; (3) EAMLS especially performs well on a) the new complicated model and b) large-scale problems. 
So why is EAMLS effective? Through combining MLS with EA and using l3-value to guide the population size to grow gradually, EAMLS performs a full local search while performing a global search, maintains good population diversity, as well as speeds up the convergence.

Our algorithm EAMLS performes well on large-scale problem instances of both m=2 and m=${\textstyle\sum_{j\in J}}X_j$ models, and its advantage will become more apparent as the problem scale increases. However, the larger the problem, the greater the number of FEs needed for EAMLS to converge.

\section{Conclusion}

This paper proposes a new RFLP formulation in which the number of facilities allocated to each customer (i.e., $m$) is not fixed but varies with decision variables $\mathbf{X}$. This non-fixed allocation setting makes the model more close to scenarios in real life.

A hybrid evolutionary algorithm EAMLS (which can also be viewed as a memetic algorithm) is proposed to solve the model. Combining a memorable local search method and EA, EAMLS performs well on the new complicated model and large-scale problems considered in this paper, and its advantage will become more obvious as the problem scale increases.
Besides, a convergence metric l3-value is proposed to analyze the algorithm's convergence speed and exam the evolutionary process.

Finally, we explore the large-scale problems of the two models. Under what conditions is a problem a large-scale problem? It is related to the model and whether the problem can be solved by the exact algorithm efficiently. For the m=2 model which allocates a fixed number of facilities to each customer as in the existing research, we solve large-scale problem instances (600-node) whose scale is much larger than other literature. For the new complicated $m=\sum_{j \in J}X_j$ model, 100-node instances can be treated as large-scale problems because the exact algorithm or optimization solver cannot solve them effectively. And our algorithm EAMLS has good performance on large-scale problems considered in this paper.

In the future, the model which integrates various factors should be studied, and more complicated FLPs, such as dynamic FLP and FLP under uncertain environments, should be focused. Furthermore, effective meta-heuristic algorithms for large-scale problems should be studied as well.

%
%
%
%

\end{document}